\documentclass{article}
\usepackage{spconf,amsmath,graphicx}
\usepackage{float}
\usepackage{tabularx}
\usepackage{adjustbox}
\usepackage{xcolor}

\title{GENERATING NATURAL QUESTIONS FROM IMAGES FOR MULTIMODAL ASSISTANTS}
\name{Alkesh Patel, Akanksha Bindal, Hadas Kotek, Christopher Klein, Jason Williams}
\address{Apple, Cupertino, CA, USA}
\begin{document}
\ninept
\maketitle
\begin{abstract}
Generating natural, diverse, and meaningful questions from images is an essential task for multimodal assistants as it confirms whether they have understood the object and scene in the images properly. The research in visual question answering (VQA) and visual question generation (VQG) is a great step. However, this research does not capture questions that a visually-abled person would ask multimodal assistants. Recently published datasets such as KB-VQA, FVQA, and OK-VQA try to collect questions that look for external knowledge which makes them appropriate for multimodal assistants. However, they still contain many obvious and common-sense questions that humans would not usually ask a digital assistant. In this paper, we provide a new benchmark dataset that contains questions generated by human annotators keeping in mind what they would ask multimodal digital assistants. Large scale annotations for several hundred thousand images are expensive and time-consuming, so we also present an effective way of automatically generating questions from unseen images. In this paper, we present an approach for generating diverse and meaningful questions that consider image content and metadata of image (e.g., location, associated keyword). We evaluate our approach using standard evaluation metrics such as BLEU, METEOR, ROUGE, and CIDEr to show the relevance of generated questions with human-provided questions. We also measure the diversity of generated questions using generative strength and inventiveness metrics. We report new state-of-the-art results on the public and our datasets.
\end{abstract}
\begin{keywords}
Multimodal assistant, computer vision, visual question generation, long-short-term memory 
\end{keywords}
\section{Introduction}
\label{sec:intro}

Current voice-based digital assistants do not take visual input into account, so there is no easy way to understand what users would ask if these assistants were able to see the real world.  One way to mimic the real scenarios is to show human annotators real-world images and ask them to note natural questions they would ask a multimodal digital assistant.  

The field of Visual Question Answering (VQA) has made incredible strides in recent years, including a large number of standard VQA datasets \cite{Jain_CVPR2017, Mostafazadeh_ACL2016, Zhang_IJCAI2016}. However, current VQA datasets \cite{Gao_NeurIPS2015, Malinowski_ICCV2015, Ren_NeurIPS2015} are focused on recognition, and most questions are about simple counting, colors, and other visual detection tasks, so these datasets do not require much association with external knowledge. The most challenging and exciting questions people ask digital assistants generally require knowing more than what the question entails or what information is contained in the images.

\begin{figure}
\begin{adjustbox}{width=\columnwidth,center}
  \includegraphics{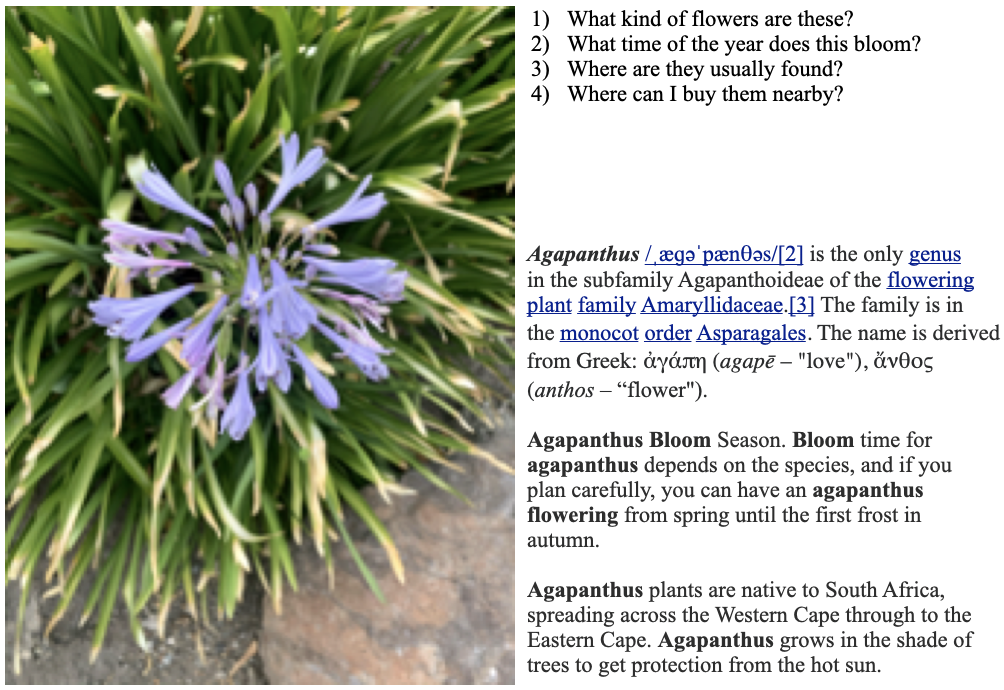}
\end{adjustbox}
  \caption{List of potential natural questions a user might ask the digital assistant by looking at the scene, Wikipedia excerpts that answer the first question, some web search results that contain answers to the second and third question. The answer to the 4th question could be a list of places where `Agapanthus’ flowers are sold.}
  \label{fig:flowers}
\end{figure}

Imagine a user comes across a flower shown in Fig. \ref{fig:flowers} for the first time. The information in the image does not say anything about where this kind of flower is grown, what time of the year they bloom, or where they can be bought. Thus, the image is not complete for answering these kinds of questions.  To answer the questions listed in Fig. \ref{fig:flowers}, we need to link the image content to external knowledge sources, such as the excerpts taken from Wikipedia, or a summary snippet from a relevant web search result.

\begin{figure}
\begin{adjustbox}{width=\columnwidth,center}
  \includegraphics{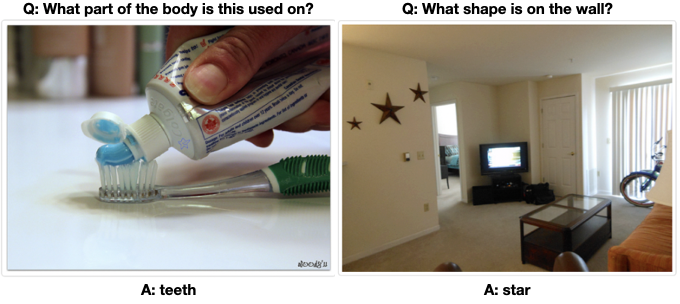}
\end{adjustbox}
  \caption{Some selected questions from the images provided in OK-VQA. The questions shown along with images may not require help from digital assistant for a visually abled person as the answers seem obvious.}
  \label{fig:dataset_diff}
\end{figure}

More recent research tries to address the shortcomings of existing VQA datasets by incorporating structured knowledge bases \cite{Marino_2019, Narasimhan_NeurIPS2018, Wang_IJCAI2017, Wang_IEEE2016, Wu_CVPR2016}  into VQA datasets. The OK-VQA dataset \cite{Marino_2019} goes a step further and also includes questions that need to reason over unstructured knowledge. In all of these datasets, questions are designed so that the answer cannot be obtained only by looking at the image.  We carefully examined the OK-VQA dataset as it is close to our work.  We make two main observations about its incompatibility for multimodal assistants: i) the image types in OK-VQA datasets are often not appropriate for inspiring any meaningful questions for the digital assistant, ii) the OK-VQA dataset has many obvious or common-sense questions for its images, as shown in Fig. \ref{fig:dataset_diff}, which are not challenging enough to ask a digital assistant.  This paper introduces an effective dataset of natural questions that are more suitable for multimodal assistant use cases.

Once we have a relevant dataset, we build the image captioning style model to generate questions from the given image. Learning to ask meaningful questions by looking at an image is an important task in NLP and vision as it demonstrates the capabilities of machine to understand the scene. Such ability can be an integral component of any digital assistant, either to engage the user to proactively start a conversation, elicit task-specific information, or suggest the question when the user cannot formulate his/her needs in natural language. We found no previous approaches providing insight into how different image and text encoding schemes can impact the quality of generated questions for a given image. We experiment with various image/text encoding schemes and decoding schemes that generate meaningful and diverse questions from the given image and associated metadata. We compare our results with previous state-of-the-art techniques on standard VQG-COCO and VQG-Flickr datasets and also provide the benchmark results on our dataset.

Our contributions in this paper are multi-fold: i) we introduce visual questions dataset that is relevant for multimodal digital assistants, ii) provide a comparison of various image and text encoding schemes and their impact on question generation, iii) build a state-of-the-art natural question generation model that takes an image and its metadata and outputs diverse and meaningful questions.

\section{Data Collection Methodology}
\label{sec:format}
\begin{table}
\centering
\label{tab:data_statistics}
\begin{adjustbox}{width=\columnwidth,center}
\begin{tabular}{| c | c | c |  p{1.5cm} | p{4cm} | p{1.2cm} | }
\hline
\textbf{Datasets} & \textbf{\# Questions}        & \textbf{\# Images }      &\textbf{Image source}   & \textbf{Goal}  & \textbf{Avg. Qn. length}    \\ 
\hline
KB-VQA  & 2402  & 700 & COCO & Visual reasoning with given KB & 6.8\\
\hline
FVQA  & 5826  & 2190 & COCO + ImageNet &  Visual reasoning with given KB & 9.5 \\
\hline
OK-VQA  & 14055 & 14031   & COCO  &  Visual reasoning with open knowledge & 8.1  \\
\hline
Ours (VQG-Apple)   & 132214  & 12006 & Flickr &  Visual reasoning with open knowledge & 7.7 \\ \cline{1-6}

\end{tabular}
\end{adjustbox}
\caption{Comparison of various visual QA datasets}
\end{table}

\begin{figure}
\begin{adjustbox}{width=\columnwidth,center}
  \includegraphics{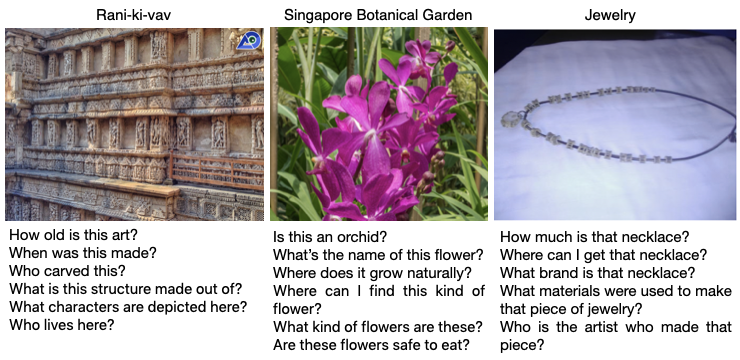}
 \end{adjustbox}
  \caption{Some examples of the questions that annotators provided from sampled images}
  \label{fig:data_collection}
\end{figure}

The image source of the OK-VQA is the COCO image dataset, which is mainly object-centric. We created a more generic dataset from carefully sampled Flickr images. \cite{Mostafazadeh_ACL2016} collected questions from images; however, their task was to generate a natural question that can potentially engage a human in starting a conversation. We adopt a similar methodology, but our guidelines emphasize asking questions about an image that one might ask a digital assistant. Our prompt shows an image and a corresponding keyword/phrase that potentially describes the image. We collected images related to products, arts, plants, flowers, animals, sculptures, places of interest, foreign text, and scenarios where people frequently seek help from a digital assistant. We sample the images by using search key terms that describe these pre-defined image categories. Since, during the search, we often get images that are not relevant to search key terms, we ask annotators first to decide if a given image is relevant for a given keyword. If they mark them as relevant, the next step is to define at least three questions that they would like to ask a digital assistant by looking at the image and metadata (e.g., location, image caption). Fig. \ref{fig:data_collection} shows some examples of the questions collected from annotators using our guidelines. Table 1 shows the salient characteristics of various knowledge-based VQA datasets and compares them with ours. As one can see, our dataset has a significantly larger number of questions from a diverse set of images from Flickr.

\section{Models}
\label{sec:pagestyle}

\begin{figure}
\begin{adjustbox}{width=\columnwidth,center}
  \includegraphics{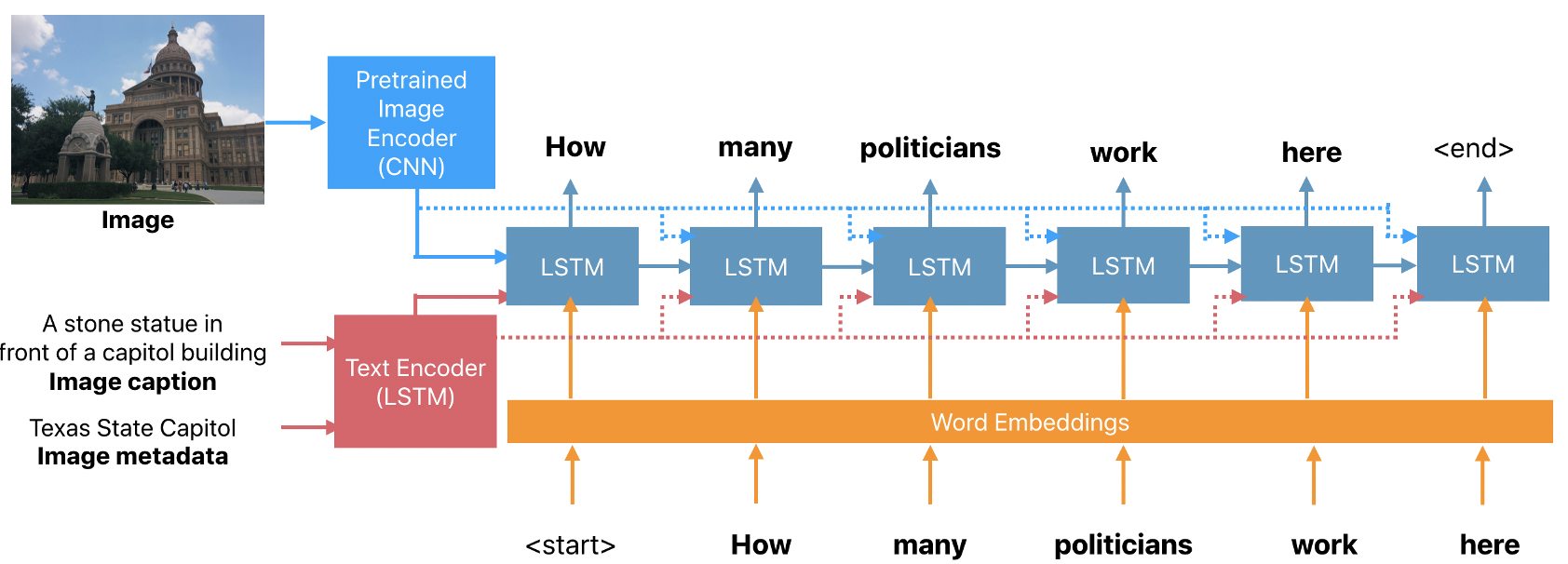}
 \end{adjustbox}
  \caption{Model architecture}
  \label{fig:model_architecture}

\end{figure}

We use a model architecture similar to image captioning \cite{Vinyals_CVPR2015}.  However, we have additional input for representing image metadata. The image metadata can be in the form of text such as a short caption describing the image, search tags/keywords associated with the image, and location in the image.  Fig. \ref{fig:model_architecture} shows the higher-level architecture of the overall model.  In computer vision literature, several pre-trained CNN models such as VGGNet \cite{Simonyan_ICLR2015}, ResNet \cite{He_CVPR2016}, MobileNet \cite{Howard_2017}, and DenseNet \cite{Huang_CVPR2017} are popular for encoding an image.  We use the image vector derived from the last layer of these CNNs to represent the image. We need to represent two types of text inputs fed into the model: i) image metadata/keywords ii) question words/text fed into encoder/decoder in every timestamp.  We can represent text using word embeddings like GloVe \cite{Penninton_EMNLP2014} or sentence embeddings derived from pre-trained networks such as ELMo \cite{Shahbaz_2019} or transformer networks like BERT \cite{Devlin_2018}.  In the default settings, we use greedy decoding scheme.  However, we additionally experiment with simple beam search with beam size of 5 and diverse beam search similar to \cite{Vijaykumar_2016} that promotes diversity in the generated questions without compromising the questions' naturalness.

In our diverse beam search decoding scheme, we first generate question tokens with a beam size of 5 until a minimum number of timestamps, T (we chose T=3). After timestamp T, we start clustering the questions generated so far across beams using token-level text similarity.  We randomly pick the cluster head from the formed clusters and keep generating question tokens.  We keep track of questions that have hit the  $\langle end \rangle$ token and add them to the result list.

\section{Experiments and Results}
\label{sec:typestyle}
\subsection{Experimental Setup}

\emph{VQG dataset}: The Visual Question Generation dataset has images from MSCOCO, Flickr, and Bing.  Each of these sets contains roughly 5000 images and five questions per image. It contains natural and engaging questions (not necessarily related to multimodal assistant use cases) about images. We noticed that in the Bing search images’ set, most of the image links were broken. So, we do not consider them in our experiments.

\noindent \emph{VQG-Apple-Flickr dataset}: We created this dataset using the data collection methodology described in Section 2.  Due to customized image selection and annotation guidelines, our images as well as corresponding questions are more tailored to multimodal assistant use cases.  There were about 12K images in our dataset. Each image has an average of nine questions, which amounts to more than 100k questions. We split the dataset into training (60\%), validation (20\%), and test (20\%) sets based on the number of images.

Since experimenting with an exhaustive list of image representation strategies is beyond this paper's scope, we used VGG19, MobileNet(v2), and DenseNet, which are pre-trained on the Imagenet dataset for our experiments to see their impact on generated questions. For representing text, we picked one static word embedding-based representation i.e. GloVe, and one contextual sequence embedding-based representation i.e. BERT. 
We first evaluate our proposed approach quantitatively on the standard automatic metrics i.e., BLEU, METEOR, ROUGE, and CIDEr.  Since none of these metrics truly represents the diversity of the generated question as they are designed to check whether the generated questions are similar to the ground truth questions or not, we used the generative strength and inventiveness metrics described in \cite{Jain_CVPR2017} to measure the naturality and diversity of the generated questions.  Generative strength tells us how many unique questions are generated by the model per image, while inventiveness indicates, out of all the generated questions, how many are new, i.e., never seen in the training data.

\subsection{Quantitative Results}
\begin{table}
\centering
\label{tab:quantitative_results1}

\begin{adjustbox}{width=\columnwidth,center}
\begin{tabular}{r|llll}
\multicolumn{1}{r}{\textbf{Experiments}}
& \multicolumn{1}{l}{\textbf{BLEU}}
& \multicolumn{1}{l}{\textbf{METEOR}}
& \multicolumn{1}{l}{\textbf{ROUGE}} 
& \multicolumn{1}{l}{\textbf{CIDEr}} \\ \cline{1-5}
\multicolumn{1}{r} {\emph{\textbf{VQG-COCO dataset}}} \\ 
\hline
\hline
VGG19 + GloVe & 30.0 & 17.1 & 42.4  & 25.2 \\
MobileNet + GloVe  & 36.6 & 21.7 & \textbf{47.4} & 50.3 \\
DenseNet + GloVe & \textbf{37.7} & \textbf{22.1} & 47.3 & \textbf{53.5} \\ \cline{1-5}
VGG19 + BERT & 29.2 & 18.2 & 42.7  & 30.3 \\
MobileNet + BERT & 30.3 & 18.6 & 43.1 & 36.4 \\
DenseNet + BERT & \textbf{31.8} & \textbf{19.4} & \textbf{43.8} & \textbf{37.9} \\ \cline{1-5}
\hline
\multicolumn{1}{r} {\emph{\textbf{VQG-Flickr dataset}}} \\ 
\hline
\hline
VGG19 + GloVe & 27.5 & 15.7 & 40.6  & 17.0 \\
MobileNet + GloVe &\textbf{27.7} & \textbf{15.8} & \textbf{40.6} & 16.1 \\
DenseNet + GloVe & 27.5 & 15.5 & 39.5 & \textbf{18.0} \\ \cline{1-5}
VGG19 + BERT & 22.1 & 15.4 & 38.6  & 16.1 \\
MobileNet + BERT & \textbf{25.3} & \textbf{16.2} & \textbf{39.8} & 17.7 \\
DenseNet + BERT & 24.4 & 15.4 & 37.6 & \textbf{18.5} \\ \cline{1-5}
\hline
\multicolumn{1}{r} {\emph{\textbf{VQG-Apple-Flickr dataset}}} \\ 
\hline
\hline
VGG19 + GloVe & \textbf{33.9} & \textbf{20.4} & \textbf{46.9}  & \textbf{22.9} \\
MobileNet + GloVe & 31.5 &  19.9 & 44.4 & 22.6 \\
DenseNet + GloVe & 32.6 & 20.1 & 45.5 & 21.6 \\ \cline{1-5}
VGG19 + BERT & 32.8 & \textbf{21.0} & 46.7  & 22.1 \\
MobileNet + BERT & \textbf{33.2} & 20.6 & \textbf{47.3} & 21.8 \\
DenseNet + BERT & 30.1 & 19.6 & 44.5 & \textbf{24.1} \\ \cline{1-5}
VGG19 + GloVe + Keywords & \textbf{33.9} & \textbf{20.4} & \textbf{46.9}  & 22.9 \\
MobileNet + GloVe + Keywords & 31.6 & 19.9 & 44.5 & 22.7 \\
DenseNet + GloVe + Keywords & 32.9 & 20.4 & 45.0 & \textbf{29.0} \\ \cline{1-5}

\end{tabular}
\end{adjustbox}
\caption{Comparison of various image encoding and text encoding schemes on VQG-COCO, VQG-Flickr, and VQG-Apple-Flickr datasets. The decoding strategy was fixed to \emph{Greedy} for these experiments.}
\end{table}

Table 2 shows the results of the experiments using the automatic metrics discussed previously.  We observe that in terms of the image encoding schemes, DenseNet performs better on VQG-COCO dataset while VGG19 and MobileNet perform better on Flickr images. Further analysis reveals that the VQG-COCO dataset has images with objects, while the Flickr dataset has generic images that contain natural indoor/outdoor scenes. This observation suggests that choice of image representation matters and if we know beforehand that there are recognizable objects in the image, we can choose one type of image representation over the others to get better results.  All pre-trained image representations we used in our experiments were using Imagenet weights i.e., trained on the same Imagenet training data. However, the underlying architecture of the network does seem to impact the outcome of the generated questions.

We notice that there is no clear pattern for the effect of text representation on the automatic evaluation metrics.  However, in general, GloVe embeddings worked better across the datasets.  We further notice that BERT based text encoding tends to generate more grammatical and long questions. However, they are often not grounded or relevant to the image.  The relatively low BLEU, METEOR, ROUGE, and CIDEr scores across all datasets for the BERT-based text encoding scheme confirms this phenomenon. For our VQG-Apple-Flickr dataset, we also notice that adding keywords either gave the same results or slightly improved them. 

\begin{table}
\centering
\label{tab:decoding_results}

\begin{adjustbox}{width=\columnwidth,center}
\begin{tabular}{r|llllll}
\multicolumn{1}{r}{\textbf{Experiments}}
& \multicolumn{1}{l}{\textbf{BLEU}}
& \multicolumn{1}{l}{\textbf{METEOR}}
& \multicolumn{1}{l}{\textbf{ROUGE}} 
& \multicolumn{1}{l}{\textbf{CIDEr}} 
& \multicolumn{1}{l}{\textbf{Gen. Str.}} 
& \multicolumn{1}{l}{\textbf{Inv. \%}} \\
\hline
\multicolumn{1}{r} {\emph{\textbf{VQG-COCO dataset}}} \\ \cline{1-7}
\hline
\hline
DenseNet + GloVe + GD & 31.0 & 19.2 & 41.8 & 40.1 & 0.7 & 54.7 \\
DenseNet + GloVe + BS& \textbf{37.8} & \textbf{22.3} & \textbf{47.5} & \textbf{53.7} & 3.9 & 44.4 \\
DenseNet + GloVe + DBS & 36.8 & 21.7 & 46.4 & 50.4 & \textbf{5.3} & \textbf{57.5} \\ \cline{1-7}
\hline
\multicolumn{1}{r} {\emph{\textbf{VQG-Apple-Flickr dataset}}} \\ 
\hline
\hline
VGG19 + GloVe + GD & 33.9 & 20.4 & 46.9  & 23.0 & 0.6 & 33.9 \\
VGG19 + GloVe + BS & \textbf{43.1} &  \textbf{23.6} & \textbf{55.2} & \textbf{40.1} & 2.3  & 24.5 \\
VGG19 + GloVe + DBS & 41.6 & 23.4 & 53.9 & 39.3 & \textbf{3.2} & \textbf{35.4} \\ \cline{1-7}

\end{tabular}
\end{adjustbox}
\caption{Comparison of various decoding schemes such as greedy(GD), simple beam search(BS) (k=5) and diverse beam search (DBS) on VQG-COCO and VQG-Apple-Flickr datasets}
\end{table}

To generate more diverse questions, we experimented with a simple beam search (beam size = 5) and a diverse beam search as decoding strategies. We computed the generative strength (Gen. Str.) and inventiveness of the generated questions for the selected configurations that were giving higher overall score for the BLEU, METEOR, ROUGE, and CIDEr.  As expected, shown in Table 3, both simple and diverse beam search generate, on average, more unique questions per image. A diverse beam search also generates more innovative questions not seen in the training data, as reflected in the inventiveness (Inv. \%) column.

\subsection{Qualitative Results}


\begin{table*}
\centering
\label{tab:decoding_results}
\begin{adjustbox}{width=1\textwidth}
\begin{tabular}{r|llll}
& \includegraphics[width=0.3\textwidth]{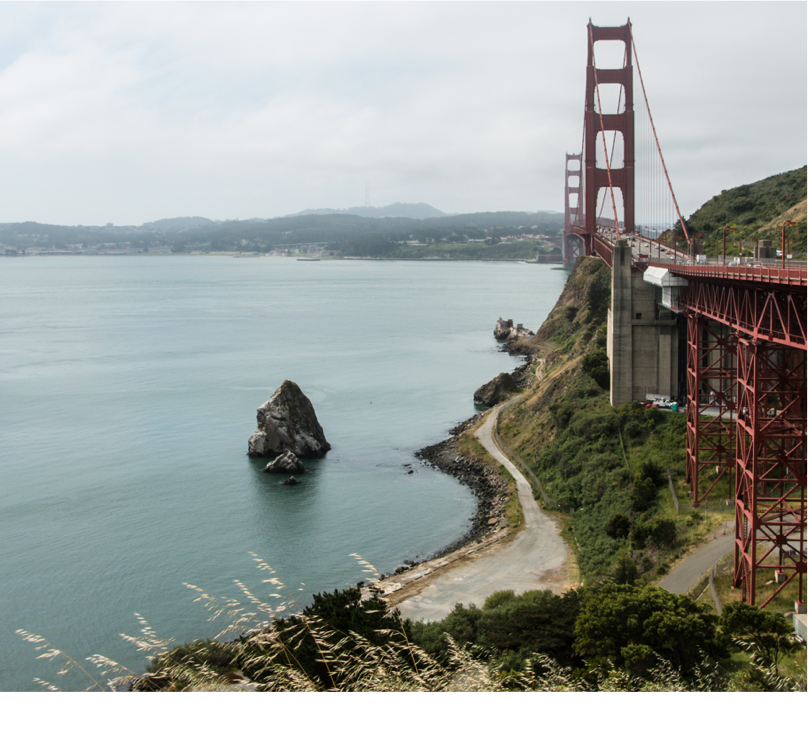} & \includegraphics[width=0.3\textwidth]{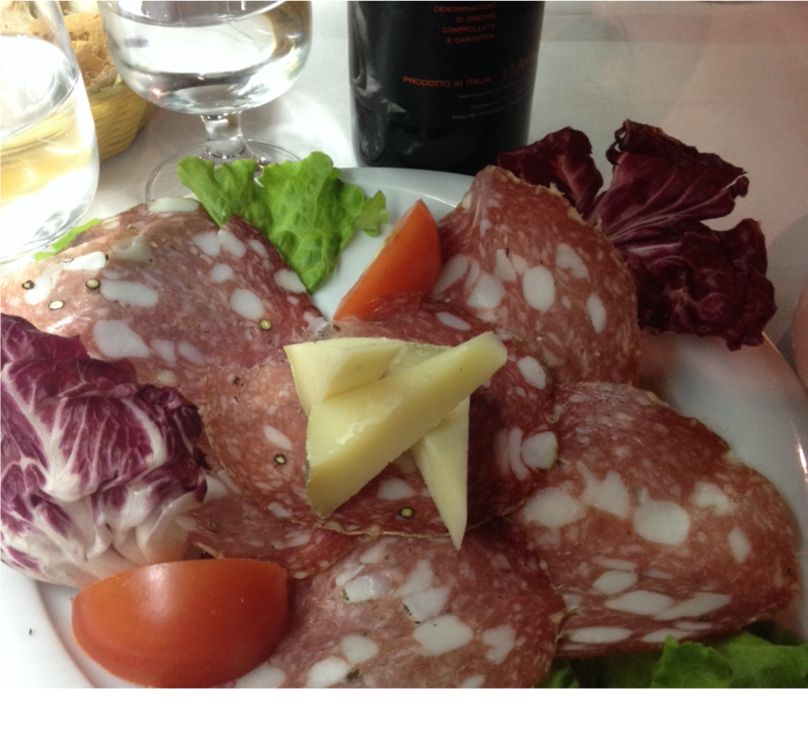} & \includegraphics[width=0.3\textwidth]{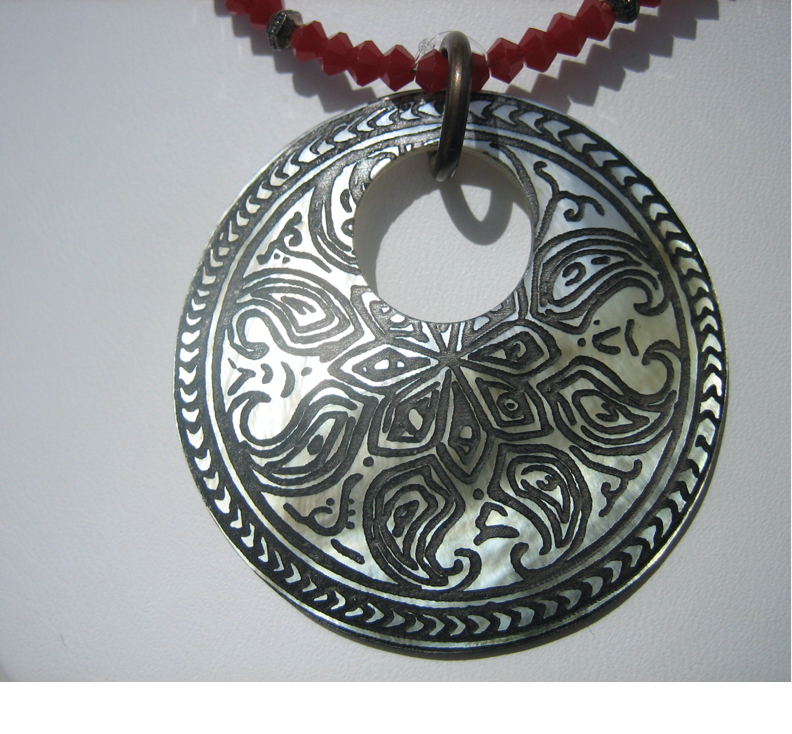} & \includegraphics[width=0.3\textwidth]{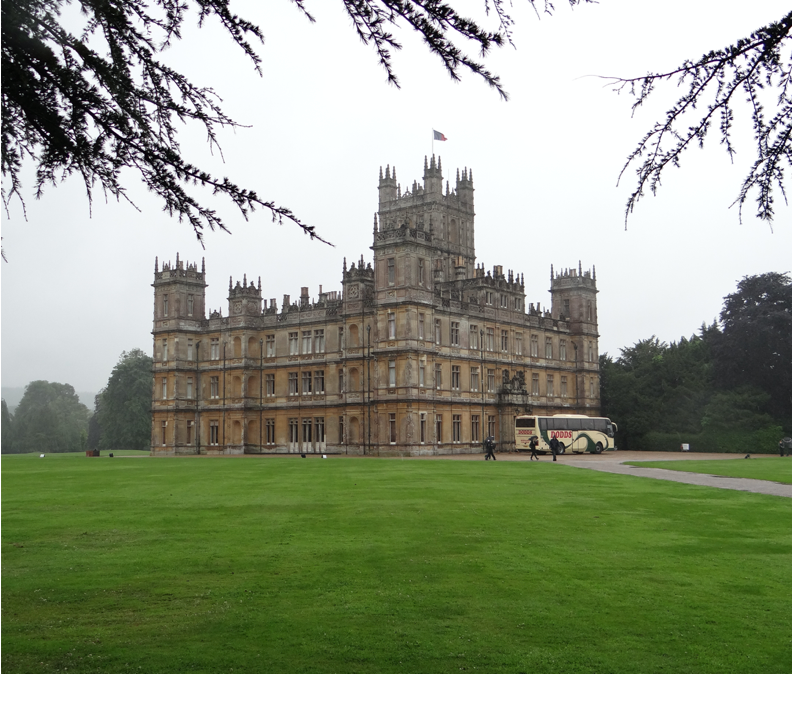} \\
& (a) & (b) & (c) & (d) \\
\hline
& how high is it & what kind of meat is that & who designed this & when was it built   \\
& how long is it & what foods are on this plate & where can i buy this & who lives there \\
& when was it built & where can i get cold cuts like this near me & is this for sale near me & can i visit this place \\
Ground Truth & how long is the bridge & what is the recipe for this dish & is this a traditional necklace of some kind & when was this built \\
& has this bridge always been painted red & how much does it cost to make this dish & what material is this & who built it \\
& is this bridge closed today & where can i go to order this dish & what is this style of art called & who currently lives there \\
& when did they last renovate this bridge & where was this purchased & how much is this necklace &  \\
& when was the bridge built & how much does this meal cost & &  \\ \cline{1-5}
\hline
\hline
VGG19 + GloVe + GD & what is the temperature of this place & what kind of meat is that & when was it built & who built this  \\
MobileNet + GloVe + GD & what body of water is that &  what kind of food is that & what kind of design is this & how old is this castle  \\
DenseNet + GloVe + GD & what kind of trees are those& what other of animals is this & who painted this & what material is the statue made of  \\ \cline{1-5}
\hline
\hline
VGG19 + BERT + GD & what is the name of this building & what kind of penguin is that & what is this made of & what is the name of this building \\
MobileNet + BERT + GD & how tall is the bridge &  what kind of food is that & what kind of bird is that & how old is this place  \\
DenseNet + BERT + GD & can i swim there & \parbox{4.5cm}{what kind of food does this ride usually grow in a day} & what is the style of architecture called & what is this style of architecture  \\ \cline{1-5}

\hline
\hline
& what body of water is that & what kind of food is that & where is this & what country is this in \\
MobileNet + GloVe + BS & what body of water is this &  where can I buy this & how much does this cost & how old is this castle \\
 &  {\color {blue}{what’s the name of that body}} &  what kind of food is this & what kind of design is this &  {\color {blue}{who owns that castle}}  \\
 & what’s the name of that body of water & {\color {blue}{what kind of dish is that}} & {\color {red} {is this in the male}} & how old is that castle  \\ \cline{2-5}

& what body of water is that & what kind of food is that & how much does this cost & how old is this castle \\
MobileNet + GloVe + DBS & what body of  water is this & where can I buy this  & what kind of design is this & how old is that castle  \\
 & {\color {red}{what’s the name of that body of}} &  {\color{red} {what kind of food is that at} } & what kind of design is that & when was it built  \\
 & {\color {blue}{what’s the name of that body}} & {\color {red} {what kind of food is that on}} & {\color {blue} {how much does it cost to make}} & {\color{blue} {how old is this castle building}}  \\ \cline{1-5}

\end{tabular}
\end{adjustbox}
\caption{Qualitative analysis of generated questions from various image encoding, text encoding, and decoding schemes on VQG-Apple-Flickr dataset. Both blue and red-colored questions are new questions that were not seen in the training set.}
\end{table*}

We perform a qualitative analysis of generated questions in randomly selected images.  Table 4 shows the ground truth questions from human annotators and questions generated from various strategies discussed in this paper. If we compare the generated questions in second row of Table 4, we notice that VGG19 \& MobileNet generate better questions, such as \emph{``what is the temperature of this place," ``when was it built," ``what body of water is that," ``what kind of food is that," ``what kind of design is this," ``how old is this castle,"} which were not part of the ground truth.  On the other hand, DenseNet produced questions such as \emph{``what kind of trees are those," ``what other of animal is this," ``what material is the statue made of,"} which are either not relevant to the image or grammatically incorrect.  This observation indicates that DenseNet, which is trained for an object detection/recognition task, may not be suitable for unbounded Flickr images.

As shown in third row, when we use BERT as the text encoding, even VGG19 and MobileNet generate questions that are completely irrelevant to the image. For example, \emph{``what is the name of the building,"} when there is no building in image (a), \emph{``what kind of penguin is that,"} when there is no penguin in image (b), \emph{``what kind of bird is that,"} when there is no bird in image (c). However, as mentioned previously, BERT-based questions were grammatically correct and relatively long.

We show the top-4 questions generated by the corresponding decoding schemes (MobileNet image encoding is picked just for relative comparison) in fourth row. A simple beam search (k=5) and diverse beam search can generate many new and diverse questions that are not only different from human-annotated ground truth questions but also not seen in the entire training set (as depicted in blue and red color). Blue-colored questions indicate that newly generated questions are acceptable for practical purposes in digital assistants.  Red-colored questions indicate that newly generated questions do not make sense for the image or not acceptable for digital assistants. Diverse beam search generates more and innovative questions such as \emph{``what’s the name of that body," ``how much does it cost to make," and ``how old is the castle building."} However, it also generates some ungrammatical quality questions (shown in red).  Instead, the beam search generates less but relatively good quality questions with some drop in generative strength and inventiveness scores.

\subsection{Comparative Results}
Table 5 shows the comparison of similar work in visual question generation literature on the VQG-COCO dataset. Our best performing configuration, i.e., DenseNet, GloVe, and beam search, produced better results by changing the image encoding scheme without making significant model architecture changes.
\begin{table}
\centering
\label{tab:comparative_results}
\begin{adjustbox}{width=\columnwidth,center}
\begin{tabular}{r|ll}
\multicolumn{1}{r}{\textbf{Approach (VQG-COCO dataset)}}
& \multicolumn{1}{l}{\textbf{BLEU}}
& \multicolumn{1}{l}{\textbf{METEOR}} \\
\hline
\hline
Mostafazadh et al. \cite{Mostafazadeh_ACL2016} & 19.2 & 19.7 \\
Jain et al. \cite{Jain_CVPR2017} & 35.6 & 19.9  \\
Ours (Best configuration) & \textbf{37.8} & \textbf{22.3}\\ \cline{1-3}
\end{tabular}
\end{adjustbox}
\caption{Comparison of similar work in Visual Question Generation}
\end{table}

\section{Conclusion and Future Work}
\label{sec:majhead}

In this paper, we presented a new visual question generation dataset that is more suitable for multimodal assistants. The number of questions is about 3.8 times more than the OK-VQA dataset, which is closest to our work. We plan to make our dataset publicly available for research use. We also presented model to automatically generate questions from images.  We experimented with various image and text encoding schemes as well as decoding schemes to generate more diverse and meaningful questions. We also compared the results of our best performing configuration with other state-of-the-art systems on standard dataset and found that choice of image encoding does matter.  In future, we would like to experiment with visio-linguistic embeddings that can be derived from recent work in multimodal transformers, such as ViLBERT\cite{Lu_NeurIPS2019}, Unicoder-VL\cite{Li_2019}, and VL-BERT \cite{Weijie_2020}.


\bibliographystyle{IEEEbib}

\end{document}